\documentclass[11pt]{article}
\usepackage{coling2016}
\usepackage{times}
\usepackage{url}
\usepackage{graphicx}
\usepackage{latexsym}
\usepackage{multirow}
\usepackage{epstopdf}
\usepackage{amssymb}
\usepackage{booktabs}
\usepackage{multirow}
\usepackage{amsmath}

\title{Bidirectional LSTM-CRF for Clinical Concept Extraction}

\author{Raghavendra  Chalapathy \\ University of Sydney\\Capital Markets CRC\\
	rcha9612@uni.sydney.edu.au
	\And  
	Ehsan Zare Borzeshi  \\ Capital Markets CRC\\ 
	ezborzeshi@cmcrc.com
	\And
	Massimo Piccardi \\University of Technology Sydney \\ 
	Massimo.Piccardi@uts.edu.au
}

\begin{document}
\maketitle
\begin{abstract}
Automated extraction of concepts from patient clinical records is an essential facilitator of clinical research. For this reason, the 2010 i2b2/VA Natural Language Processing Challenges for Clinical Records introduced a concept extraction task aimed at identifying and classifying concepts into predefined categories (i.e., treatments, tests and problems). State-of-the-art concept extraction approaches heavily rely on handcrafted features and domain-specific resources which are hard to collect and define. For this reason, this paper proposes an alternative, streamlined approach: a recurrent neural network (the bidirectional LSTM with CRF decoding) initialized with general-purpose, off-the-shelf word embeddings. The experimental results achieved on the 2010 i2b2/VA reference corpora using the proposed framework outperform all recent methods and ranks closely to the best submission from the original 2010 i2b2/VA challenge.
\end{abstract}

\vspace{0.5cm}
\section{Introduction}
\label{intro}

Patient clinical records typically contain longitudinal data about patients' health status, diseases, conducted tests and response to treatments. Analysing such information can prove of immense value not only for clinical practice, but also for the organisation and management of healthcare services. \textit{Concept extraction} (CE) aims to identify mentions to medical concepts such as problems, test and treatments in clinical  records (e.g., discharge summaries and progress reports) and classify them into predefined categories. The concepts in  clinical records are often expressed with unstructured, ``free'' text, making their automatic extraction a challenging task for clinical Natural Language Processing (NLP) systems. 
Traditional approaches have extensively relied on rule-based systems and lexicons to recognise the concepts of interest. Typically, the concepts represent drug names, anatomical nomenclature and other specialized names and phrases which are not part of everyday vocabularies. For instance, ``resp status'' should be interpreted as ``response status''. Such use of abbreviated phrases and acronyms is very common within the medical community, with many abbreviations having a specific meaning that differ from that of other lexicons. Dictionary-based systems perform concept extraction by looking up terms on medical ontologies such as the Unified Medical Language System (UMLS) ~\cite{kipper2008system}. Intrinsically, dictionary- and rule-based systems are laborious to implement and inflexible to new cases and misspellings~\cite{liu2015drug}. Although these systems can achieve high precision, they tend to suffer from low recall (i.e., they may miss a significant number of concepts). To overcome these limitations, various machine learning approaches have been proposed (e.g., conditional random fields (CRFs), maximum-entropy classifiers and support vector machines) to simultaneously exploit the textual and contextual information while reducing the reliance on lexicon lookup~\cite{lafferty2001conditional,berger1996maximum,joachims1998text}. State-of-the-art machine learning approaches usually follow a two-step process of \textit{feature engineering} and \textit{classification}. The feature engineering task is, in its own right, very laborious and demanding on expert knowledge, and it can become the bottleneck of the overall approach. For this reason, this paper proposes a highly streamlined alternative: to employ a contemporary neural network - the bidirectional LSTM-CRF - initialized with general-purpose, off-the-shelf word embeddings such as GloVe ~\cite{Pennington:14} and Word2Vec~\cite{Mikolov:13}. The experimental results over the authoritative 2010 i2b2/VA benchmark show that the proposed approach outperforms all recent approaches and ranks closely to the best from the literature. 

\begin{table*}[]
	\centering
	\scalebox{0.95}{
	\begin{tabular}{|c|c|c|c|c|c|c|c|c|c|c|}
		\hline \bf Sentence & \textit{His}& \textit{HCT} & \textit{had}& \textit{dropped} & \textit{from} &\textit{36.7} &\textit{despite} &\textit{2U} &\textit{PRBC} &\textit{and}  \\ 
		\hline \textbf{Concept class}& \textit{O}& \textit{B-test}& \textit{O} & \textit{O}& \textit{O} & \textit{O} & \textit{ O} & \textit{B-treatment} & \textit{I-treatment} & \textit{O} \\ 			
		\hline
	\end{tabular}}
	\caption{Example sentence in a concept extraction task. The concept classes are represented in the standard in/out/begin (IOB) format.}
	\label{table1}
\end{table*}

\section{Related Work}
\label{relatedworks}
Most of the research to date has framed CE as a specialized case of named-entity recognition (NER) and employed a number of supervised and semi-supervised machine learning algorithms with domain-dependent attributes and text features~\cite{uzuner20112010}. Hybrid models obtained by cascading CRF and SVM classifiers along with several pattern-matching rules have shown to produce effective results~\cite{boagcliner}. Moreover, \cite{fu2014improving} have given evidence to the importance of including preprocessing steps such as truecasing and annotation combination. The system that has reported the highest accuracy on the 2010 i2b2/VA concept extraction benchmark is based on unsupervised feature representations obtained by Brown clustering and a hidden semi-Markov model as classifier~\cite{de2011machine}. However, the use of a ``hard'' clustering technique such as Brown clustering is not suitable for capturing multiple relations between the words and the concepts. For this reason, Jonnalagadda et al. \cite{jonnalagadda2012enhancing} demonstrated that a random indexing model with distributed word representations can improve clinical concept extraction. Moreover, Wu et al. \cite{wu2015study} have jointly used word embeddings derived from the entire English Wikipedia \cite{collobert2011natural} and binarized word embeddings derived from domain-specific corpora (e.g. the MIMIC-II corpus \cite{MIMIC2}). In the broader field of machine learning, the recent years have witnessed a proliferation of deep neural networks, with outstanding results in tasks as diverse as visual, speech and named-entity recognition~\cite{hinton2012deep,krizhevsky2012imagenet,lample2016neural}. One of the main advantages of neural networks over traditional approaches is that they can learn the feature representations automatically from the data, thus avoiding the expensive feature-engineering stage. Given the promising performance of deep neural networks and the recent success of unsupervised word embeddings in general NLP tasks \cite{Pennington:14,Mikolov:13,lebret2013word}, this paper sets to explore the use of a state-of-the-art deep sequential model initialized with general-purpose word embeddings for a task of clinical concept extraction.

%
%

\section{ The Proposed Approach}
CE can be formulated as a joint segmentation and classification task over a predefined set of classes. As an example, consider the input sentence provided in Table~\ref{table1}. The notation follows the widely adopted in/out/begin (IOB) entity representation with, in this instance, \textit{HCT} as the test and \textit{2U PRBC} as the treatment. In this paper, we approach the CE task by the bidirectional LSTM-CRF framework where each word in the input sentence is first mapped to either a random vector or a vector from a word embedding. We therefore provide a brief description of both word embeddings and the model hereafter.  

Word embeddings are vector representations of natural language words that aim to preserve the semantic and syntactic similarities between them. The vector representations can be generated by either count-based approaches such as Hellinger-PCA~\cite{lebret2013word} or trained models such as Word2Vec (including skip-grams and continuous-bag-of-words) and GloVe trained over large, unsupervised corpora of general-nature documents. In its embedded representation, each word in a text is represented by a real-valued vector, $x$, of arbitrary dimensionality, $d$.

Recurrent neural networks (RNNs) are a family of neural networks that operate on sequential data. They take as input a sequence of vectors ($x_1,x_2, ...,x_n$) and output a sequence of class posterior probabilities, ($y_1,y_2,...,y_n$). An intermediate layer of hidden nodes, ($h_1,h_2,...,h_n$), is also part of the model. In an RNN, the value of the hidden node at time $t$, $h_{t}$, depends on both the current input, $x_t$, and the previous hidden node, $h_{t-1}$. This recurrent connection from the past timeframe enables a form of short-term memory and makes the RNNs suitable for the prediction of sequences. Formally, the value of a hidden node is described as:

\begin{equation}
h_t = f(U \bullet x_t + V \bullet h_{t-1} )
\end{equation}
\vspace{0.2cm}

\noindent where $U$ and $V$ are trained weight matrices between the input and the hidden layer, and between the past and current hidden layers, respectively. Function $f(\cdot)$ is the sigmoid function, $f(x)={1}/{1+e^{-x}}$, that adds non-linearity to the layer. Eventually, $h(t)$ is input into the output layer and convolved with the output weight matrix, $W$:

\begin{equation}
\label{eq4}
y_t = g(W \bullet h_t), \hspace{0.03in} \text{with} \hspace{0.03in}  g(z_{m}) = \frac{e^{z_{m}}}{\Sigma _{k=1}^Ke^{z_{k}} } 
\end{equation}
\vspace{0.2cm}

Eventually, the output is normalized by a multi-class logistic function, $g(\cdot)$, to become a proper probability over the class set. Therefore, the output dimensionality is equal to the number of concept classes. Although an RNN can, in theory, learn long-term dependencies, in practice it tends to be biased towards its most recent inputs. For this reason, the Long Short-Term Memory (LSTM) network incorporates an additional ``gated'' memory cell that can store long-range dependencies~\cite{hochreiter1997long}. In its bidirectional version, the LSTM computes both a forward, $\overrightarrow{h_t}$, and a backward, $\overleftarrow{h_t}$, hidden representation at each timeframe $t$. The final representation is created by concatenating them as $h_t = [\overrightarrow{h_t}$;$\overleftarrow{h_t}]$. In all these networks, the hidden layer can be regarded as an implicit, learned feature that enables concept prediction. A further improvement to this model is provided by performing joint decoding of the entire input sequence in a Viterbi-style manner using a CRF ~\cite{lafferty2001conditional} as the final output layer. The resulting network is commonly referred to as the \textit{bidirectional LSTM-CRF} \cite{lample2016neural}.

\section{Experiments}

\subsection{Dataset}
\label{sec:length}
\begin{table*}[]	
	\centering
	\scalebox{1.1}{
	\begin{tabular}{|c|c|c|}
		\hline
		& Training set & Test set\\		
		\hline
		notes &$170$ &$256$  \\
		sentences &$16315$&$27626$\\ 
		\hline		
		problem  & $7073$   & $12592$  \\		
		test&  $4608$ & $9225$\\		
		treatment& $4844$ & $9344$\\
		\hline
	\end{tabular}}
	\caption{Statistics of the training and test data sets used for the 2010-i2b2/VA concept extraction.}
	\label{table2} 
\end{table*}

The 2010 i2b2/VA Natural Language Processing Challenges for Clinical Records include a concept extraction task focused on the extraction of medical concepts from patient reports. For the challenge, a total of 394 concept-annotated reports for training, 477 for testing, and 877 unannotated reports were de-identified and released to the participants alongside a data use agreement~\cite{uzuner20112010}. However, part of this data set is no longer being distributed due to restrictions later introduced by the Institutional Review Board (IRB). Thus, Table~\ref{table2} summarizes the basic statistics of the training and test data sets which are currently publicly available and that we have used in our experiments.

\subsection{Evaluation Methodology}
\label{sec:length}
Our models have been blindly evaluated on the 2010 i2b2/VA CE test data using a strict evaluation criterion requiring the predicted concepts to exactly match the annotated concepts in terms of both boundary and class. To facilitate the replication of our experimental results, we have used a publicly-available library for the implementation of the LSTM (i.e. the Theano neural network toolkit \cite{bergstra2010theano}) and we publicly release our code\footnote{https://github.com/raghavchalapathy/Bidirectional-LSTM-CRF-for-Clinical-Concept-Extraction}. We have split the training set into two parts (sized at approximately 70\% and 30\%, respectively), using the first for training and the second for selection of the hyper-parameters (``validation'')~\cite{bergstra2012random}.The hyper-parameters include the embedding dimension, $d$, chosen over $\{50, 100, 300, 500\}$, and two additional parameters, the learning and drop-out rates, that were sampled from a uniform distribution in the range $[0.05, 0.1]$. All weight matrices were randomly initialized from the uniform distribution within range $[-1, 1]$. The word embeddings were either initialized randomly in the same way or fetched from Word2Vec and GloVe~\cite{w2vcode,glovecode}. Approximately $25\%$ of the tokens were alphanumeric, abbreviated or domain-specific strings that were not available as pre-trained embeddings and were always randomly initialized. Early stopping of training was set to $50$ epochs to mollify over-fitting, and the model that gave the best performance on the validation set was retained. The accuracy is reported in terms of micro-average F$_1$ score computed using the CoNLL score function~\cite{Nadeau:07}.

\subsection{Results and Analysis}
Table \ref{table3} shows the performance comparison between state-of-the-art CE systems and the proposed bidirectional LSTM-CRF with different initialization strategies. As a first note, the bidirectional LSTM-CRF initialized with GloVe outperforms all recent approaches (2012-2015). On the other hand, the best submission from the 2010 i2b2/VA challenge~\cite{de2011machine} still outperforms our approach. However, based on the description provided in \cite{uzuner20112010}, these results are not directly comparable since the experiments in~\cite{de2011machine,jonnalagadda2012enhancing} had used the original dataset which has a significantly larger number of training samples. Using general-purpose, pre-trained embeddings improves the F$_1$ score by over 5 percentage points over a random initialization. In general, the results achieved with the proposed approach are close and in many cases above the results achieved by systems based on hand-engineered features.

\begin{table*}[]	
	\centering
	\scalebox{1.1}{
	\begin{tabular}{|c|c|c|c|}
		\hline
		Methods & Precision & Recall & F$_1$ Score \\
		\hline
		Hidden semi-Markov Model \cite{de2011machine} &$86.88$ &$83.64$ & $85.23$ \\
		Distributonal Semantics CRF \cite{jonnalagadda2012enhancing} &$85.60$&$82.00$ &$83.70$  \\	
		\hline
		Binarized Neural Embedding CRF \cite{wu2015study}&$85.10$&$80.60$ & $82.80$ \\
		CliNER \cite{boagcliner}&$79.50$&$81.20$ & $80.00$ \\
		Truecasing CRFSuite \cite{fu2014improving}&$80.83$&$7 1.47$ & $75.86$ \\		
		\hline
		Random - Bidirectional LSTM-CRF &$81.06$ &$75.40$ & $78.13$ \\
		Word2Vec - Bidirectional LSTM-CRF &$82.61$ &$80.03$ & $81.30$ \\
		GloVe - Bidirectional LSTM-CRF &$84.36$ &$83.41$ & $83.88$ \\
		\hline	
	\end{tabular}}
	\caption{Performance comparison between the bidirectional LSTM-CRF (bottom three lines) and state-of-the-art systems (top five lines) over the 2010 i2b2/VA concept extraction task.}
	\label{table3}
\end{table*}

\label{sec:length}

\section*{Conclusion}
This paper has explored the effectiveness of the contemporary bidirectional LSTM-CRF for clinical concept extraction. The most appealing feature of this approach is its ability to provide end-to-end recognition using general-purpose, off-the-shelf word embeddings, thus sparing effort from time-consuming feature construction. The experimental results over the authoritative 2010 i2b2/VA reference corpora look promising, with the bidirectional LSTM-CRF outperforming all recent approaches and ranking closely to the best submission from the original 2010 i2b2/VA challenge. A potential way to further improve its performance would be to explore the use of unsupervised word embeddings trained from domain-specific resources such as the MIMIC-III corpora \cite{MIMIC}. 

\bibliographystyle{acl}
\bibliography{coling2016}

\end{document}